\documentclass[final,5p,times,twocolumn,authoryear]{elsarticle}

\usepackage{amssymb}
\usepackage{lipsum}
\usepackage{url}
\usepackage{amsthm}
\usepackage{amsmath}
\usepackage{amsfonts}
\usepackage{algorithmic}
\usepackage{algorithm}
\usepackage{array}
\usepackage{textcomp}
\usepackage{stfloats}
\usepackage{verbatim}
\usepackage{graphicx}
\usepackage{cite}
\usepackage{color}
\usepackage{multirow}
\usepackage{colortbl}
\usepackage{xcolor}

\usepackage{soul}

\usepackage[colorlinks=true, linkcolor=blue, citecolor=blue, urlcolor=blue]{hyperref}

\begin{document}

\begin{frontmatter}

\title{Plug-and-Play DISep: Separating Dense Instances for Scene-to-Pixel Weakly-Supervised Change Detection in High-Resolution Remote Sensing Images}

\author[first,second]{Zhenghui Zhao}
\author[first,second]{Chen Wu\corref{cor1}}
\author[fourth]{Lixiang Ru}
\author[third]{Di Wang}
\author[fifth,sixth]{Hongruixuan Chen}
\author[third]{Cuiqun Chen}
\cortext[cor1]{Corresponding author: Chen Wu}
\fntext[fn]{This work was supported by the National Key Research and Development Program of China under Grant 2022YFB3903300, and partly by the National Natural Science Foundation of China under Grant T2122014, 61971317.}

\affiliation[first]{organization={State Key Laboratory of Information Engineering in Surveying, Mapping and Remote Sensing, Wuhan University}, addressline={Wuhan},postcode={430072}, country={China}}
\affiliation[second]{organization={Institute of Artificial Intelligence, Wuhan University}, addressline={Wuhan},postcode={430072}, country={China}}
\affiliation[third]{organization={School of Computer Science, Wuhan University}, addressline={Wuhan},postcode={430072}, country={China}}
\affiliation[fourth]{organization={Ant Group}, 
            addressline={Hangzhou},   
            postcode={310013},
            country={China}}
\affiliation[fifth]{organization={Graduate School of Frontier Sciences, University of Tokyo}, addressline={Chiba},   
            postcode={277-8561},
            country={Japan}}
\affiliation[sixth]{organization={Institute of Geodesy and Photogrammetry, ETH Zürich}, addressline={Zürich},   
            postcode={999034},
            country={Switzerland}}

\begin{abstract}
Change Detection (CD) focuses on identifying specific pixel-level landscape changes in multi-temporal remote sensing images. The process of obtaining pixel-level annotations for CD is generally both time-consuming and labor-intensive. Faced with this annotation challenge, there has been a growing interest in research on Weakly-Supervised Change Detection (WSCD). WSCD aims to detect pixel-level changes using only scene-level (i.e., image-level) change labels, thereby offering a more cost-effective approach. Despite considerable efforts to precisely locate changed regions, existing WSCD methods often encounter the problem of "instance lumping" under scene-level supervision, particularly in scenarios with a dense distribution of changed instances (i.e., changed objects). In these scenarios, unchanged pixels between changed instances are also mistakenly identified as changed, causing multiple changes to be mistakenly viewed as one. In practical applications, this issue prevents the accurate quantification of the number of changes. To address this issue, we propose a Dense Instance Separation (DISep) method as a plug-and-play solution, refining pixel features from a unified instance perspective under scene-level supervision. Specifically, our DISep comprises a three-step iterative training process: 1) Instance Localization: We locate instance candidate regions for changed pixels using high-pass class activation maps. 2) Instance Retrieval: We identify and group these changed pixels into different instance IDs through connectivity searching. Then, based on the assigned instance IDs, we extract corresponding pixel-level features on a per-instance basis. 3) Instance Separation: We introduce a separation loss to enforce intra-instance pixel consistency in the embedding space, thereby ensuring separable instance feature representations. The proposed DISep adds only minimal training cost and no inference cost. It can be seamlessly integrated to enhance existing WSCD methods. We achieve state-of-the-art performance by enhancing {three Transformer-based and four ConvNet-based methods} on the LEVIR-CD, WHU-CD, DSIFN-CD, SYSU-CD, and CDD datasets. Additionally, our DISep can be used to improve fully-supervised change detection methods. Code is available at \url{https://github.com/zhenghuizhao/Plug-and-Play-DISep-for-Change-Detection}.

\end{abstract}

\begin{keyword}
Deep Learning \sep Change Detection \sep Weakly-Supervised Learning \sep High-Resolution Remote Sensing

\end{keyword}

\end{frontmatter}




\section{Introduction}
\label{sec:intro}
Change Detection (CD) is a cornerstone task in remote sensing that aims to identify specific pixel-level landscape changes within multi-temporal remote sensing images. The significance of CD spans various domains, including urban planning \citep{lee2021local, Singh2020,10565926}, ecological monitoring \citep{wu9,song2014remote,Fatima2021}, and disaster assessment \citep{xu2019building,chen2023exchange,zheng2021building}. In the current era of deep learning, dominant CD methods rely on a fully-supervised paradigm that requires costly pixel-level annotations. However, the recent global expansion of high-resolution satellites has highlighted the need to balance annotation costs with effective data utilization. In this context, scene-to-pixel (i.e., image-level) Weakly-Supervised Change Detection (WSCD) presents an attractive trade-off between performance and annotation costs. WSCD only labels pairs of bi-temporal scene images as either ‘changed’ or ‘unchanged,’ thereby inspiring numerous significant works due to its potential to greatly reduce annotation efforts \citep{wu2023,Huang2023,wang2023cs}. In essence, the WSCD paradigm only indicates whether the scenes have changed, requiring the identification of specific pixel-level change localizations.

\begin{figure}
    \centering
    \includegraphics[scale=0.27]{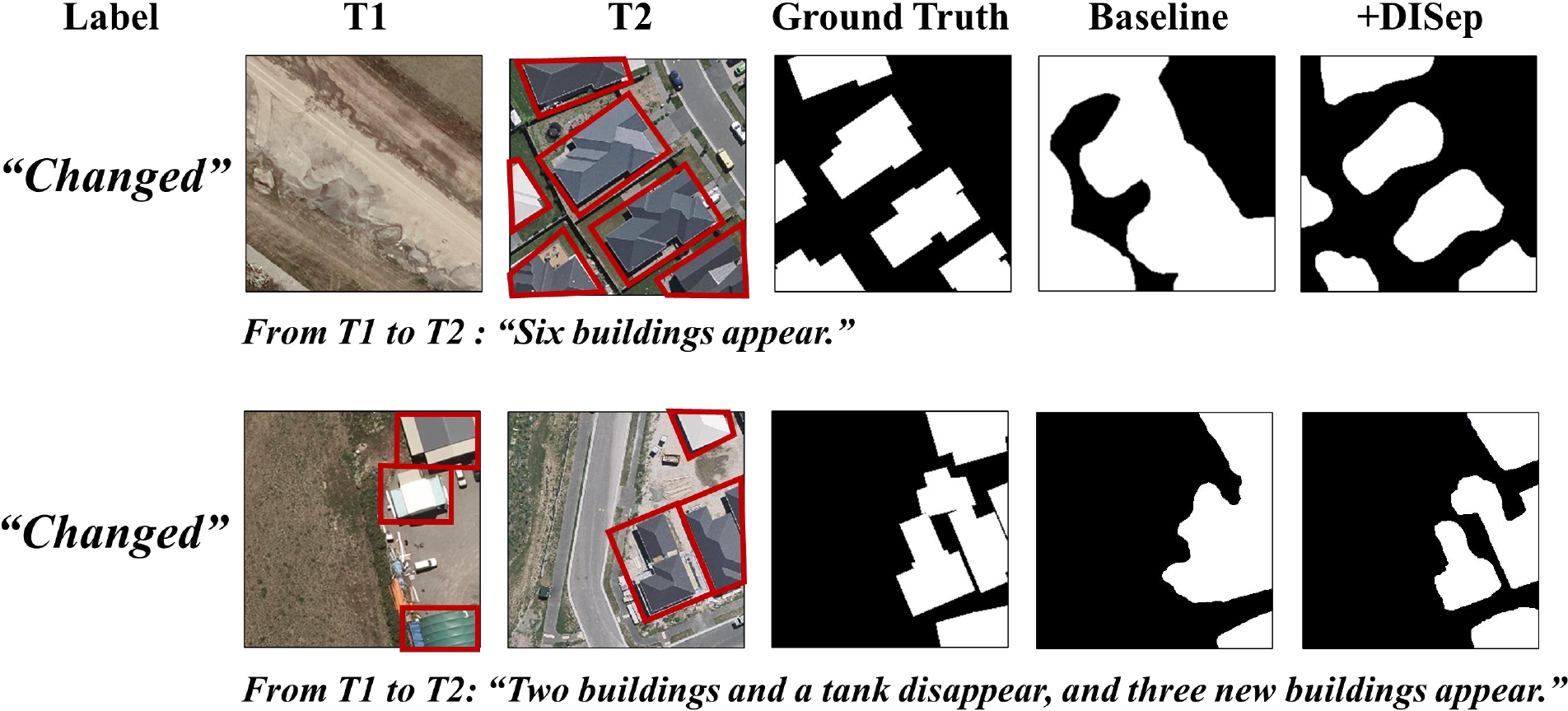}
    \caption{Motivation for our DISep. We show the change predictions in dense instance scenarios, comparing the baseline method (e.g., TransWCD) against the enhanced method with our DISep. In the baseline method, changed instances tend to merge together, resulting in \textit{instance lumping}. Our DISep successfully separates these merging instances.}
    \label{introduction1}
\end{figure}

\begin{figure*}[ht]
    \centering
    \includegraphics[scale=0.56]{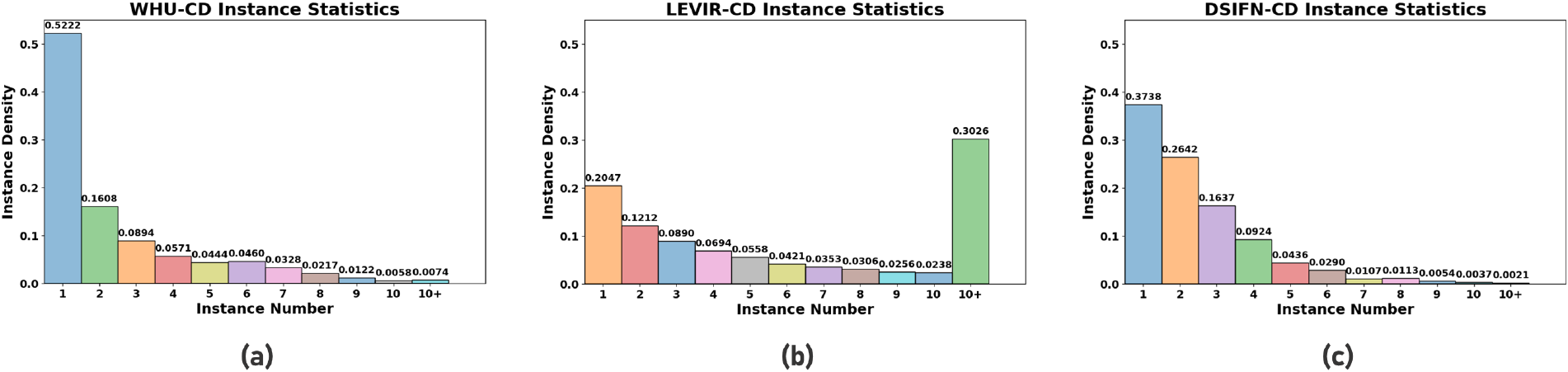}
    \caption{Prevalence of dense instance distribution in change detection. We present statistics on the instance distribution within the WHU-CD and LEVIR-CD datasets. (a) WHU-CD: 47.77\% of image pairs contain multiple instances. (b) LEVIR-CD: It exhibits a higher density. 79.53\% of the examples have multiple instances, and 30.26\% contain more than 10 instances. (c) DSIFN-CD: It exhibits 62.62\% of the paired image examples involving multiple instances.}
    \label{introduction2}
\end{figure*}

The current workflow for WSCD generally follows these four steps: 1) Inputting bi-temporal paired images and their scene-level binary labels (changed or unchanged) into classification models. 2) Using a scene-level classification loss (i.e., cross-entropy loss) to constrain the classification models. 3) Extracting class localization maps of changed pixels from the classification models. 4) Obtaining pixel-level change predictions by retaining salient pixels in class localization maps. The feasibility of achieving pixel-level predictions from scene-level supervision arises from the deep models' tendency to focus on regions that are most significant to the classification decision \citep{zhou2016learning}.

Despite this structured workflow, existing WSCD approaches still only achieve coarse pixel-level change localization under scene-level supervision due to the large gap between scene-level supervision and pixel-level predictions. In scenes with densely packed changed instances, these rough localizations often result in unchanged pixels between changed instances being mistaken for changed ones. Consequently, this gives rise to the issue of instance lumping, as shown in Figure \ref{introduction1}. In the first example, the six buildings that appear are merged into a single object of change in the predictions. Similarly, in the second example, buildings that disappear at time T1 and those that appear at time T2 are combined, resulting in them being mistakenly identified as a single change.

This instance lumping hinders accurate quantification of the number of changes in practical applications. For example, in damage assessment following natural disasters, accurately identifying damaged structures is crucial. If multiple damaged targets are considered as one, the extent of the damage may be underestimated, affecting rescue and reconstruction efforts. Furthermore, such dense instance distribution, which is prevalent in remote-sensing change detection, as demonstrated by the statistics in Figure \ref{introduction2}. The WHU-CD dataset shows that 47.77\% of its images contain multiple instances, the DSIFN-CD dataset has 62.62\% of images with multiple instances, and the LEVIR-CD dataset presents an even higher multi-instance ratio of 79.53\%.

To address the challenge of instance lumping, we propose a \underline{D}ense \underline{I}nstance \underline{Sep}aration (DISep). Our DISep comprises three iterative procedures: instance localization, retrieval, and separation. First, we extract high-confidence changed pixels and create instance localization masks for locating changed instances from Class Activation Maps (CAMs). Here, without pixel-level labels, a high-pass threshold is applied to make a trade-off between the instance sampling size and statistical Type-I errors of changed pixels. Second, within these localization masks, we conduct instance retrieval, assigning different IDs to each isolated changed region, thereby creating the instance identity masks for these changed pixels. Finally, we employ a separation loss to guide an intra-instance pixel online clustering, resulting in more instance-discriminative pixel feature embeddings. Additionally, for class balance, unchanged background regions are treated as individual instances, promoting a comprehensive method for instance separation.

Our DISep is based on the rationale that pixels within the same instance should exhibit greater feature consistency, without compromising changed or unchanged intra-class similarity. By incorporating instance identity information into pixels under scene-level supervision, DISep significantly improves WSCD performance in scenarios with multiple instances. Its minimal training cost and no extra inference overhead make DISep an ideal plug-and-play module for existing frameworks. We evaluate DISep on top of seven ConvNet-based and Transformer-based WSCD methods to demonstrate its effectiveness and versatility. Experiments show that DISep consistently boosts the performance of existing WSCD methods, setting new state-of-the-art records on the WHU-CD, LEVIR-CD, DSIFN-CD, SYSU-CD, and CDD datasets. Additionally, our DISep can be used to enhance the performance of fully-supervised change detection.


Our contributions can be summarized as follows:
\begin{itemize}
    \item {We propose DISep, a novel approach for WSCD that incorporates instance identity information into pixels, tackling the issue of \textit{instance lumping} in scenarios with dense changes.}
    \item {We introduce a separation loss to enhance the intra-instance pixel feature consistency, refining pixel features from a unified instance perspective under scene-level supervision.}
    \item {Our DISep can be integrated as a plug-and-play module with existing WSCD methods, significantly improving the accurate quantification of changed objects using only scene-level annotations.}
\end{itemize}

\section{Related Work}\label{sec:related}
\subsection{Weakly-Supervised Change Detection}
The field of Weakly-Supervised Change Detection (WSCD) gains prominence in 2017 with the pioneering work of Khan \textit{et al.} \citep{khan2017}, introducing the first deep learning-based approach to WSCD, employing Conditional Random Fields (CRFs) along with conventional neural networks (ConvNets). After this, WCDNet \citep{ander2020} incorporates a novel remapping block and refines change predictions using CRFs with recurrent neural networks. These early works primarily utilize CRF-based methods, leveraging pixel correlations to refine predictions in post-event (T2) images, often overlooking the changes disappearing from pre-event (T1) images. Pixel correlation-based techniques, including CRFs \citep{sutton2010introduction}, cannot effectively handle changes across multiple temporal images. Subsequent research explores alternative technologies to enhance WSCD.

Kalita \textit{et al.} \citep{kal2021} integrate Principal Component Analysis (PCA) and the K-means algorithm with a custom Siamese convolutional network. Daudt \textit{et al.} conduct extensive research in weakly supervised change detection, initially proposing a method that combines guided anisotropic diffusion and iterative learning \citep{daudt2019guided}. Building upon this foundation, they explore the use of guided anisotropic diffusion for weakly supervised change detection, demonstrating robustness and applicability across diverse remote sensing datasets \citep{daudt2023weakly}. BGMix \citep{Huang2023} introduces background-mixed augmentation for augmented and real data. FCD-GAN \citep{wu2023} develops a generative adversarial network-based model combining multiple supervised learning methods for change detection, including weakly supervised learning with fully convolutional networks.

More recently, Zhao \textit{et al.} \citep{transwcd} establish an efficient framework called TransWCD. Additionally, they develop TransWCD-DL, which includes a label-gated constraint and a prior-based decoder to rectify inconsistencies between predictions and labels. MS-Former \citep{MSFormer} proposes a memory-supported transformer model designed for patch-level weakly supervised change detection. This approach utilizes more granular sub-image annotations, known as patch-level annotations, which provide a finer level of detail compared to traditional image-level labels. MSCAM \citep{MSCAM} introduces a multi-scale approach combined with adaptive online noise correction techniques to enhance high-resolution change detection of built-up areas, providing a robust framework to manage noise and improve detection accuracy in urban environments. CS-WSCDNet \citep{wang2023cs} employs large-scale visual models that generate pixel-level pseudo labels from the Segment Anything Model (SAM) \citep{kirillov2023segment}. Despite these advancements, most methods still overlook the challenge of dense instance scenarios, leading to suboptimal performance in such contexts.

\subsection{Instance Estimation with CAMs}
Weakly-Supervised Change Detection (WSCD) typically derives initial pixel-level predictions from a classification model. This process involves retaining strongly salient pixels within class localization maps, highlighting discriminative regions crucial for classification decisions under scene-level supervision \citep{zhou2016learning}. Currently, Class Activation Maps (CAMs) are the preferred method to acquire class localization maps in WSCD \citep{wang2023cs}. In WSCD tasks, CAMs can achieve more complete change localizations, as the bi-temporal paired image inputs facilitate identification.

To our knowledge, instance estimation using scene-level labels has yet to be explored in change detection. However, CAMs have proven helpful in capturing instance cues in other weakly-supervised dense prediction tasks, such as weakly-supervised instance segmentation in computer vision \citep{zhou2016learning,oquab2015object,kim2022beyond,ponttuset2017multiscale}. Oquab \textit{et al.} \citep{oquab2015object} explore weakly-supervised object localization using CNNs, showing that CNNs trained with image-level labels can localize objects by identifying the most discriminative regions. MCG \citep{ponttuset2017multiscale} presents a multiscale combinatorial grouping approach for image segmentation and object proposal generation, combining hierarchical image segmentation with combinatorial grouping to generate high-quality object proposals. Kim \textit{et al.} \citep{kim2022beyond} propose a framework for weakly-supervised instance segmentation using semantic knowledge transfer and self-refinement, refining instance segmentation predictions.

These weakly-supervised instance segmentation methods are not directly applicable to WSCD due to the unique instance-dense distribution characteristic of remote sensing change detection tasks. Additionally, edge-based instance refinement methods \citep{ma2023eatder, li2020improving, zhang2023aernet} are ineffective for WSCD, as changes usually involve the overlay of two images. Without pixel-level labels, it is challenging to determine which image contains the change in WSCD, let alone edge information.

However, CAM-based weakly-supervised instance segmentation methods demonstrate that it is feasible to implement instance localization using CAMs for WSCD. Specifically for WSCD, we incorporate instance identity information into a plug-and-play solution, avoiding the use of more complex parametric modules commonly found in instance segmentation methods.

\subsection{Similarity Metric Optimization}
Metric learning, focusing on optimizing a distance metric to bring similar samples closer and push dissimilar ones apart, has emerged as a crucial approach in enhancing classification tasks by improving intra-class compactness and inter-class separability \citep{sohn2016improved}. It is widely considered an effective complement to cross-entropy loss. Metric learning mitigates the issues of cross-entropy loss, such as sensitivity to adversarial examples \citep{elsayed2018large, nar2019cross}, the tendency to ignore inductive biases \citep{mettes2019hyperspherical, mitchell1980need}, and the production of poor classification margins, which can significantly impact classification performance \citep{cao2019learning}.

Some fully-supervised change detection (FSCD) methods have explored enhancing change detection performance from a metric perspective. Touati \textit{et al.} \citep{Touati2020} and Guo \textit{et al.} \citep{Guo2021} employ feature distance constraints to improve the similarity between cross-temporal pixels at corresponding positions directly. Besides, there is a growing trend in using contrastive \citep{SCPFCD, zhang2022beyond, zhan2017change} or triplet \citep{Shi2022} metric learning for FSCD, especially for creating positive and negative sample pairs of changed and unchanged pixels. These FSCD works mainly employ a cross-temporal paired metric approach within and between classes, aiming to decrease the feature distance among unchanged pixel pairs while increasing it for changed ones, guided by pixel-level ground truths.

However, even setting aside the reliance on pixel-level labels, these FSCD metric approaches still fail to address the issue of instance lumping. This is because these approaches focus only on the feature similarity between cross-temporal pixel pairs, ignoring the cross-spatial feature consistency. In contrast, we introduce a novel metric method, DISep, designed specifically for WSCD. Our DISep utilizes intra-instance pixel-to-centroid online clustering across spatial pixel embeddings, improving the cross-spatial instance consistency of pixels in feature space.

\section{Dense Instance Separation}
\label{sec:method}
\begin{figure*}
	\centering
	\includegraphics[scale=0.56]{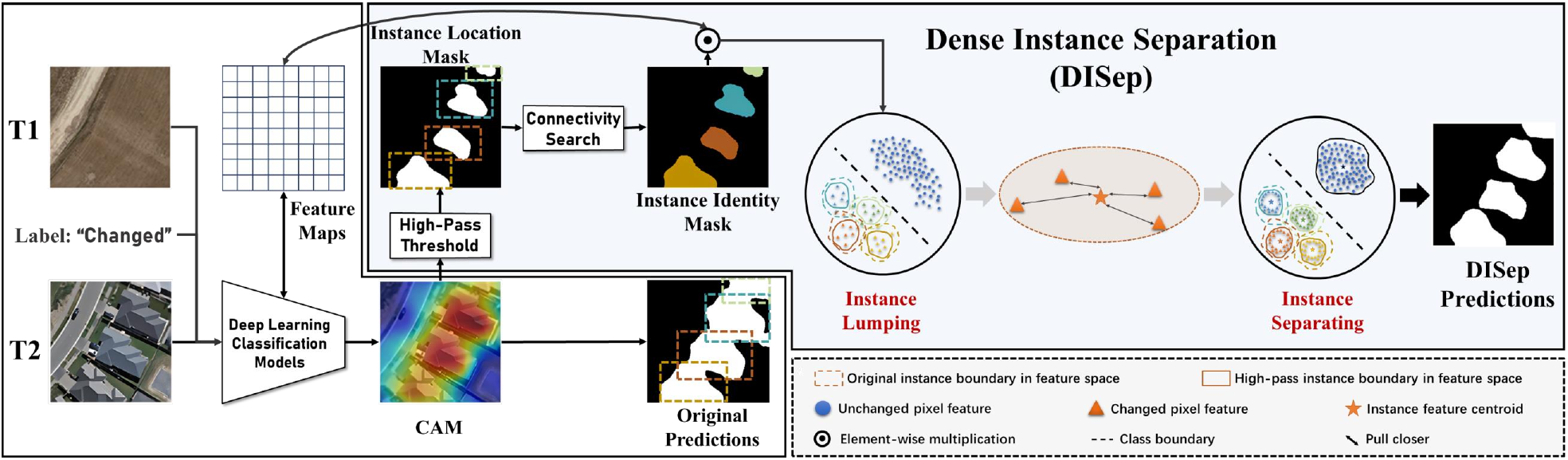}
	\caption{Overview of our DISep. First, we obtain instance localization masks from the CAM using a high-pass threshold. Then, we implement instance retrieval through connectivity search and acquire instance identity masks. Finally, we employ a separation loss to guide an intra-instance pixel online clustering in feature space, according to the instance identity masks. Note that the high-pass threshold is used for optimization to select reliable samples, not for CAM prediction generation.} 
	\label{method1}
 \vspace{1cm} 
\end{figure*}

Our DISep is the first work to address the issue of instance lumping commonly in scenarios with dense changed instances. As shown in Figure \ref{method1}, DISep is designed to seamlessly integrate with existing models as a plug-and-play module, guiding the training phase without adding computational overhead during the inference phase.

\subsection{Preliminary}
We begin with the preliminary step of generating class activation maps (CAMs) from a scene-level supervised change classification model.

Let $X=\{(x_{t1}^n,x_{t2}^n)\}_{n=1}^N$ represent the bi-temporal paired image space, and let $y_{cls} = \{changed (1), unchanged (0)\}$ represent the scene-level change label space, and define a WSCD training dataset $X_{train} = \{(x_{t1}^n, x_{t2}^n, y_{cls}^n)\}_{n=1}^{N}$. Then, we randomly sample paired images $(x_{t1}, x_{t2})$ of size $HW \times 3$, along with the corresponding scene-level label $y_{cls}$, and input them into a change classification model that includes a bi-temporal difference module processing paired images. 

The change scene classification model outputs the last-layer feature maps (i.e., change classification logits) $\mathcal{F}\in \mathbb{R}^{HW \times D}$, where $D$ denotes the channel dimension. Then, the CAMs $\mathcal{C}\in \mathbb{R}^{HW}$ are generated using the weight matrices $W \in \mathbb{R}^{D}$ from the last classification layer:
\begin{equation}
\label{cam}
\mathcal{C}=\frac{\mathcal{CAM}^{i}}{\text{Max}(\mathcal{CAM}^{i})}, \; \text{where} \; \mathcal{CAM} = \text{ReLU}(\sum_{j}^{D} \mathcal{F}^{:,j}W^{j}).
\end{equation}
where the $\text{ReLU}$ function eliminates the negative activation value and the $\text{Max}$ normalization scales the values to $[0,1]$. Subsequently, we introduce a preset CAM score to differentiate the changed and unchanged regions, yielding the change pseudo-label predictions.

\subsection{Instance Localization via High-Pass CAMs} \label{sec:method1}
In WSCD, paired input images ensure more complete change region identification from CAMs. Furthermore, peak instance responses \citep{zhou2018weakly} aid in localizing instances from CAMs under scene-level supervision. These observations help us approach instance localization using a high-pass threshold in WSCD. For sample balance, we treat background regions as individual unchanged instances.

Specifically, we introduce a high-pass threshold $T_h \in [0,1]$ to extract changed instance localizations from CAMs of the $n$-th changed image pairs labeled with $y_{cls}=1$. This threshold, set above the CAM score, helps remove false positive unchanged pixels and retain reliably changed pixels, resulting in a binary changed instance localization mask $\mathcal{M}_c \in \mathbb{R}^{H\times W}$:
\begin{equation}
\mathcal{M}_c = \begin{cases}
1, & \textit{if} \; \mathcal{C}^{i} \geq T_h \\
0, & \textit{otherwise}
\end{cases}.
\end{equation}
In the changed instance localization mask, the high-pass threshold isolates originally merged changed instances from each other. 

Concurrently, the remaining unchanged background pixels in the $n$-th changed example are grouped into a single unchanged instance. The unchanged instance localization mask $\mathcal{M}_{uc}$ is then created by preserving pixels $(i)$ where the CAM values are below a low-pass threshold $T_l$:
\begin{equation}
\mathcal{M}_{uc} = \begin{cases}
1, & \textit{if} \; \mathcal{C}^{i} \leq T_l \\
0, & \textit{otherwise}
\end{cases}.
\end{equation}
Under scene-level weak supervision, we have to set high- and low-pass thresholds, Th and Tl, to discard uncertain pixels, thus ensuring the accuracy of the selected pixels. The high- and low-pass thresholds are specifically used to select reliable samples for optimization. By eliminating these uncertain pixels, we can enhance the consistency among instances. It is important to note that, as with existing WSCD methods, we still use the CAM score to differentiate between changed and unchanged regions, yielding the change pseudo-label predictions. From the perspective of Bayesian statistics, DISep introduces additional prior information, thereby facilitating more accurate predictions of changes. This process is iterative, continuously refining the model's understanding of the relationships between instances by providing constantly refreshed priors.
 
\subsection{Instance Retrieval by Connectivity Search}  \label{sec:method2}

In this section, we identify and assign unique instance IDs to each pixel within the changed instance localization mask $\mathcal{M}_c$ using connectivity search. 

Specifically, we introduce an instance identity mask $\mathcal{M}_{id}$, initially set to all-0 and dimensionally consistent with the instance localization mask $\mathcal{M}_c$. Then, we conduct a pixel-by-pixel sequential traversal on $\mathcal{M}_c$, leveraging an 8-neighborhood connectivity search, where the changed instance localization regions are considered as foreground.

For every candidate pixel where $\mathcal{M}_c(i,j)=1$, we examine its instance identity value $\mathcal{M}_{id}(i,j)$, as well as the values of its 8 neighbors $\mathcal{N}_{id}(i, j)$, where $\mathcal{N}_{id}(i,j) = \{\mathcal{M}_{id}(i + \delta_i, j + \delta_j) \mid \delta_i, \delta_j \in \{-1, 0, 1\}, (\delta_i, \delta_j) \neq (0, 0)\}$. The entire process is detailed as follows:
\begin{figure}
	\centering
	\includegraphics[scale=0.25]{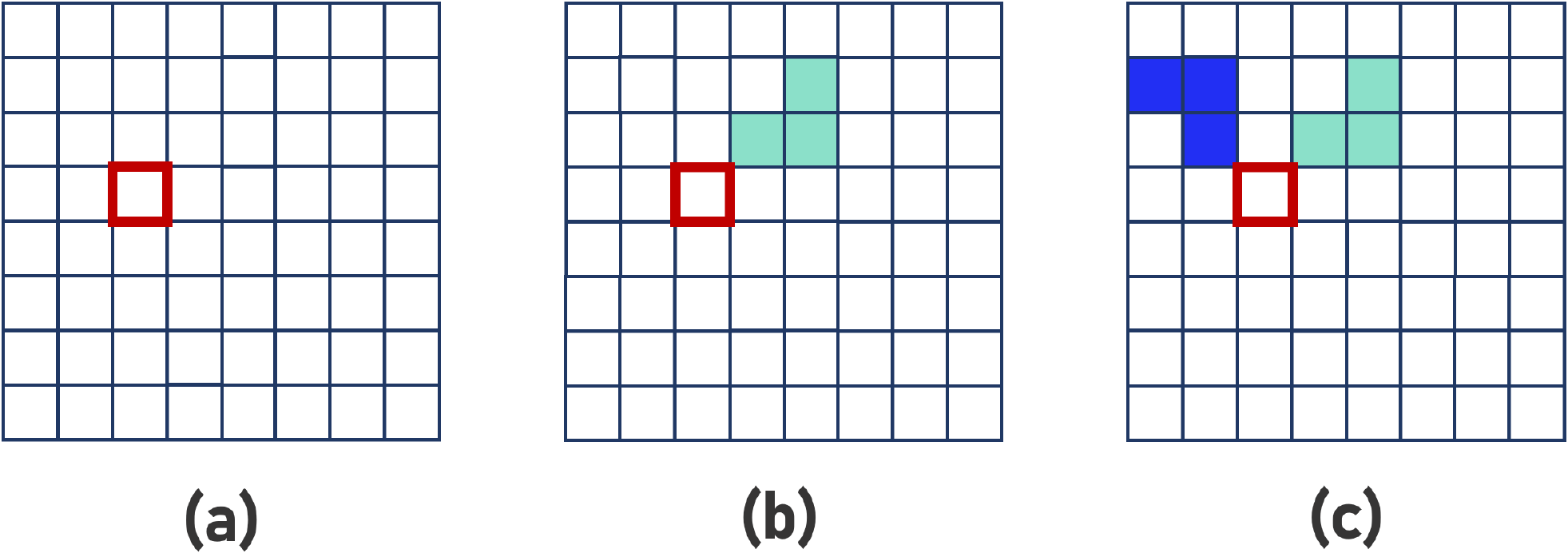}
	\caption{Three cases of changed instance retrieval. The square with a red border indicates the current pixel. (a) The pixel belongs to a new changed instance. (b) The pixel belongs to an already existing changed instance. (c) The pixel connects two existing changed instances, merging them into one.}
	\label{method2}
\end{figure}

a) If $\mathcal{N}_{id}(i, j)=\mathbf{0}$, it indicates that all the adjacent positions of pixel $(i,j)$ are background pixels, and this pixel belongs to a new changed instance. In this case, we assign an instance ID value $k$ to $\mathcal{M}_{id}(i,j)$.

b) If $\mathcal{N}_{id}(i, j) \neq \mathbf{0}$ and contains only one instance ID value $\mathcal{N}_{id}(i, j) = \{k\}$, it signifies that the adjacent positions of pixel $(i,j)$ belong to the same existing changed instance. We then update the current pixel's instance ID into $k$.

c) If $\mathcal{N}_{id}(i, j)\neq\mathbf{0}$ and contains different instance IDs $\mathcal{N}_{id}(i, j) = \{k_1, k_2, ...\}$, it suggests that pixel $(i,j)$ is connected to multiple previously isolated changed regions with different instance IDs. In such scenarios, we update all the pixels connected to $(i,j)$ into $k$, effectively merging these changed candidate instances into a single one.

These three cases are shown in Figure \ref{method2}. During this traversal, the pixels of different isolated regions on $\mathcal{M}_c$ are assigned different instance IDs $k \in \mathbb{Z}^+$. We update the corresponding pixels on the instance identity mask $\mathcal{M}_{id}$. The instance IDs $k$ indicate which specific instance each changed pixel $(i, j)$ belongs to.

Subsequently, we implement instance-level feature extraction, using the instance identity mask $\mathcal{M}_{id}$ in the output feature maps $\mathcal{F}$. Formally,
\begin{equation}
\mathcal{F}_{c_k} = \begin{cases}
\mathcal{F}^{i}, & \textit{if } \mathcal{M}_{id}^{i} = k \\
0, & \textit{otherwise}
\end{cases},
\end{equation}
where $\mathcal{F}_{c_k}$ denotes the $k$-th changed instance feature. Similarly, we employ the unchanged instance localization mask $\mathcal{M}_{uc}$ to directly extract the remaining unchanged background feature $\mathcal{F}_{uc}$ from the output feature of changed images:
\begin{equation}
\mathcal{F}_{uc} = \mathcal{F} \odot \mathcal{M}_{uc},
\end{equation}
where $\odot$ represents a pixel-by-pixel multiplication.
\subsection{Instance Separation with Intra-Instance Consistency } \label{sec:method3}
Guided by the instance identity mask, we propose a separation loss designed to conduct intra-instance online clustering, enhancing pixel-level feature consistency. The clustering anchor is defined as the instance centroids, which are the average features of the pixels within an instance. This approach avoids disturbing the intra- and inter-class similarity of changed and unchanged pixels.

Specifically, we compute the separation loss, denoted as $\mathcal{L}_{sep}$, based on three different instance-image category correlations. In WSCD, instances can be categorized as three types based on their relationship with the corresponding image cases: \textit{changed-in-changed} (i.e., changed instances in changed images), \textit{unchanged-in-changed} (i.e., unchanged background in changed images), and \textit{unchanged-in-unchanged} (i.e., entire unchanged images). The separation loss $\mathcal{L}_{sep}$ considers all three instance types to ensure sample balance. Formally, it is expressed as:
\begin{equation}
\mathcal{L}_{sep} = l_{p_c} + l_{p_{uc}} + l_{p_{u}},
\label{eq_px}
\end{equation}
where $l_{p_c}$, $l_{p_{uc}}$, and $l_{p_u}$ represent different constraint branches for the \textit{changed-in-changed}, \textit{unchanged-in-changed}, and \textit{unchanged-in-unchanged} pixels, respectively. By employing the separation loss $\mathcal{L}_{sep}$, DISep approaches WSCD from a unified instance perspective within the pixel-level embedding space, under scene-level supervision.

In the context of \textit{changed-in-changed} instances, we calculate the centroid $p_{c_k}$ of the $k$-th changed instance feature, $\mathcal{F}_{c_k}$, as follows:
\begin{equation}
p_{c_k} = \frac{1}{N_{c_k}} \sum^{HW}_{i} \mathcal{F}_{c_k}^{i},
\end{equation}
where $N_{c_k}$ denotes the count of pixels within the $k$-th changed instance, and features outside the instance's region are set to zero in the representation $\mathcal{F}_{c_k} \in \mathbb{R}^{HW \times D}$. We then promote changed intra-instance feature consistency by a pixel-to-centroid clustering constraint $l_{p_c}$:
\begin{equation}
l_{p_c} = \frac{1}{N_{c_k}} \sum_{i}^{HW} \| \mathcal{F}_{c_k}^{i} - p_{c_k} \|_2^2,
\end{equation}
utilizing the Euclidean distance $\| \cdot \|_2$. Sometimes, the centroids of some changed instances may be close in feature space, but this is not a concern.

We also introduce a constraint for the unchanged background between changed instances to ensure separation in the actual physical space. The \textit{unchanged-in-changed} constraint $l_{p_{uc}}$ is as follows:
\begin{equation}
l_{p_{uc}} = \frac{1}{N_{uc}} \sum_{i}^{HW} \| \mathcal{F}_{uc}^{i} - p_{uc} \|_2^2,
\end{equation}
where $N_{uc}$ is the pixel number of the unchanged instance, and $p_{uc}$ denotes the centroid of the unchanged-in-changed branch. Specifically, $p_{uc}$ is calculated by averaging the feature vectors of all pixels within the unchanged region, which is located between the changed instances:
\begin{equation}
p_{uc} = \frac{1}{N_{uc}} \sum_{i}^{HW} \mathcal{F}_{uc}^i.
\end{equation}
$N_{uc}$ denotes the count of pixels within the unchanged instance. $\mathcal{F}_{uc}$ represents the feature vectors within the unchanged region, with features outside the instance's region are set to zero in the representation $\mathcal{F}_{uc} \in \mathbb{R}^{HW \times D}$. Similar to the calculation of $p_{c_k}$, features outside the unchanged region are set to zero in $\mathcal{F}_{uc}$.

The \textit{unchanged-in-unchanged} branch $l_{p_{u}}$ is:
\begin{equation}
l_{p_u} = \frac{1}{HW} \sum_{i}^{HW} \| \mathcal{F}^{i} - p_{u} \|_2^2,
\end{equation}
where we calculate the clustering of all pixels in the feature map $\mathcal{F}$ towards the unchanged image center $p_u$. Specifically, $p_u$ is calculated by averaging the feature vectors of all pixels within the unchanged image:
\begin{equation}
p_u = \frac{1}{HW} \sum_{i}^{HW} \mathcal{F}^i,
\end{equation}
where $HW$ represents the total number of pixels in the unchanged image, and $\mathcal{F}$ denotes the feature vectors of all pixels.

Beyond the current intra-instance clustering, we also investigated the potential of enhancing inter-instance discriminability through centroid contrast among the three instance types. However, we observed that the addition of centroid-wise contrast leads to a degradation in performance. Further analysis revealed that including centroid-wise contrast and other typical inter-instance constraints, such as triplet and contrastive losses, adversely affected intra-class similarity. This observation suggests that this inter-instance constraint may disrupt intra-class similarity, detailed further in Sec.~\ref{sec:diff_loss}.

The overall loss function in our end-to-end pipeline consists of the separation loss and the binary cross-entropy loss, formalized as:
\begin{equation}
\mathcal{L} = \mathcal{L}_{cls} + \alpha \mathcal{L}_{sep},
\label{eq_total}
\end{equation}
where $\alpha$ is the weight factor of $\mathcal{L}_{sep}$ controlling its contribution in the end-to-end pipeline.

\section{Experiments}
\subsection{Experimental Setup}
\subsubsection{Datasets}
We conduct experiments on five publicly available change detection datasets: WHU-CD \citep{ji2018fully}, LEVIR-CD \citep{chen2020spatial}, DSIFN-CD \citep{zhang2020deeply}, SYSU-CD \citep{shi2021deeply}, and CDD \citep{shi2021deeply}. These datasets cover various scenarios, from urban development to different land-cover changes.

\begin{figure}[ht]
	\centering
	\includegraphics[width=0.42\textwidth]{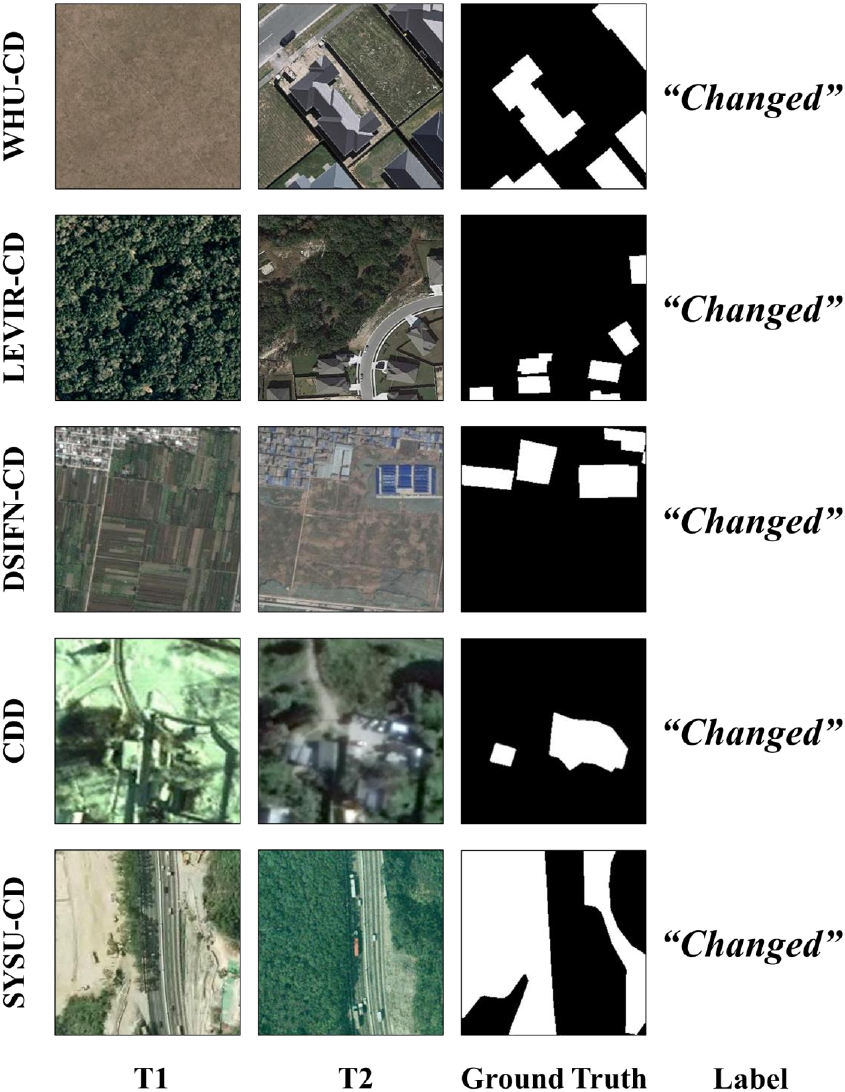}
	\caption{Examples of the WHU-CD, LEVIR-CD, DSIFN-CD, SYSU-CD, and CDD datasets.}
	\label{appendix1}
\end{figure}

The LEVIR-CD dataset is specifically designed to address the challenges in large-scale building change detection. It comprises 637 pairs of high-resolution (0.5 m) remote sensing images, each with a dimension of 1024×1024 pixels. To facilitate the training and inference process, these images are further segmented into non-overlapping patches of 256×256 pixels, yielding a comprehensive dataset of 7,120 training examples, 1,024 validation examples, and 2,048 test examples. This distribution includes 4,538 examples of change and 5,654 examples without change, offering a balanced dataset for practical model training and benchmarking.

The WHU-CD dataset focuses on urban building change detection, featuring a single pair of high-resolution (0.3 m) aerial images spanning an extensive area of 32,507×15,354 pixels, captured over a four-year interval (2012 to 2016). The dataset is meticulously processed into 256×256 pixel examples, resulting in 5,544 unchanged and 1,890 changed examples. The division of data into 60\% for training, 20\% for validation, and 20\% for testing ensures a robust framework for model assessment across different phases of learning.

The DSIFN-CD dataset broadens the scope of change detection to include a variety of land-cover objects such as roads, buildings, croplands, and water bodies, making it one of the most diverse datasets in this field. It contains six pairs of high-resolution (2 m) satellite images from six distinct cities, each showing unique environmental and developmental features. According to a standard 256×256 pixel split, the dataset is organized into 14,400 training, 1,360 validation, and 192 test examples. With 2,426 unchanged and 13,526 changed examples, it presents a challenging set for models to accurately identify a wide range of change types.

The SYSU-CD dataset captures urban changes in Hong Kong from 2007 to 2014 through 20,000 pairs of high-resolution aerial images, each 256×256 pixels. It includes changes such as new buildings, suburban sprawl, groundwork, vegetation shifts, road extensions, and offshore constructions. The dataset is split into 12,000 training pairs, 4,000 validation pairs, and 4,000 test pairs, providing a comprehensive evaluation framework for change detection algorithms in urban settings.

The CDD dataset focuses on seasonal changes within the same area, using high-resolution satellite images collected from Google Earth. These images, with resolutions ranging from 0.03 m to 1 m, are cropped to 256×256 pixels and divided into 10,000 training, 3,000 validation, and 3,000 test sets. This setup allows for the evaluation of change detection models in identifying various changes, including urban development and vegetation shifts, across different contexts.




\subsubsection{Evaluation Protocol} 
In line with established evaluation practices \citep{Huang2023}, three key metrics are commonly used: Overall Accuracy (OA), Intersection over Union (IoU), and the F1 score. 

OA measures the proportion of correctly identified pixels (changed and unchanged) to the total number of pixels, providing a general view of model accuracy. IoU, a crucial indicator for spatial overlap accuracy, calculates the intersection ratio to the union of predicted and actual change areas, directly reflecting spatial accuracy in CD. The F1 score, calculated as the harmonic mean of precision (the proportion of correctly predicted changed pixels to the total predicted changed pixels) and recall (the ratio of correctly predicted changed pixels to the total actual changes), balances the trade-off between precision's focus on minimizing false positives and recall's emphasis on reducing false negatives, making it the primary metric for evaluating change detection performance \citep{Zhang2022}.

\subsubsection{Baselines}
Our methodology is evaluated on seven notable WSCD methods to assess its effectiveness. These baseline methods include the ConvNet-based approaches of Kalita \textit{et al.} \citep{kal2021}, WCDNet \citep{ander2020}, FCD-GAN \citep{wu2023}, and BGMix \citep{Huang2023}, alongside the Transformer-based TransWCD \citep{transwcd}, MSCAM \citep{MSCAM}, and MS-Former \citep{MSFormer}. All the ConvNet-based methods utilize VGG-16 \citep{simonyan2014very} as their backbone architecture. We employ CAMs to derive class localization maps from these methods for pixel-level predictions. For Transformer-based methods like TransWCD and MSCAM, which natively integrate CAMs, we evaluate our DISep model directly without further modifications. MS-Former is reproduced with CAMs in a manner similar to the ConvNet-based methods.

\subsubsection{Experimental Settings} 
Following the default configurations of baseline models, we maintain the same optimizer, pretraining, training epochs, learning rate, and data augmentation strategies. {Bi-temporal images are processed at a resolution of $256 \times 256$ pixels}, and the CAM score is set at 0.45. The high-pass threshold for generating changed instance localization masks is 0.60, and the low-pass threshold for unchanged instances is 0.40. The weight factor of the separation loss is set to $\alpha=0.1$ in Eq. \ref{eq_total}. We train for 10,000 iterations and integrate the separation loss $\mathcal{L}_{sep}$ after 200 iterations. This phased integration is designed to leverage the early learning stages for establishing a reliable foundation of instance localization, upon which the separation loss can then be built to refine the model's ability to accurately detect changes progressively.

\subsection{Comparison to State-of-the-Arts}
\begin{table*}
\setlength{\tabcolsep}{2pt}
\renewcommand{\arraystretch}{1.2}
\centering
\caption{Quantitative improvements over baselines and comparisons with weakly-supervised change detection methods on building change detection datasets. F1 score (\%), OA (\%), and IoU (\%) are reported. Red tiny text indicates the improvements brought by our DISep method, and bold font indicates the highest performance values achieved.}
\label{WSCD_b}
\begin{tabular}{cccccccc} 
\hline\hline
\multirow{2}{*}{\textbf{Method}}                  & \multicolumn{3}{c}{\textbf{WHU-CD}}                                                                                &  & \multicolumn{3}{c}{\textbf{\textbf{LEVIR-CD}}}                                                                                                                                           \\ 
\cline{2-4}\cline{6-8}
                                                  & \textbf{F1}                          & \textbf{OA}                          & \textbf{IoU}                         &  & \textbf{F1}                          & \textbf{\textbf{OA}}                 & \textbf{\textbf{IoU}}                \\ 
\hline
Kalita \textit{et al.}                            & 31.35& 80.66& 18.27&  & 26.01                                & 84.40                                & 17.95                                \\
\rowcolor[rgb]{0.925,0.925,0.925} \textbf{+DISep} & 33.27\tiny\textcolor{red}{+1.92}& 85.90\tiny\textcolor{red}{+5.24}& 20.56\tiny\textcolor{red}{+2.29}&  & 30.82\tiny\textcolor{red}{+4.81}          & 89.37\tiny\textcolor{red}{+4.97}          & 22.07\tiny\textcolor{red}{+4.12}          \\ 
\hline
WCDNet                                            & 35.43& 80.30                                & 20.10                                &  & 32.56                                & 88.45                                & 18.79                                \\
\rowcolor[rgb]{0.925,0.925,0.925} \textbf{+DISep} & 43.32\tiny\textcolor{red}{+7.91}& 88.25\tiny\textcolor{red}{+7.95}          & 26.34\tiny\textcolor{red}{+6.24}          &  & 36.97\tiny\textcolor{red}{+4.41}          & 93.55\tiny\textcolor{red}{+5.10}          & 22.68\tiny\textcolor{red}{+3.89}          \\ 
\hline
FCD-GAN                                           & 52.44& 77.58                                & 35.32                                &  & 43.08                                & 94.49                                & 30.45                                \\
\rowcolor[rgb]{0.925,0.925,0.925} \textbf{+DISep} & 56.55\tiny\textcolor{red}{+4.11}& 84.01\tiny\textcolor{red}{+6.43}          & 39.80\tiny\textcolor{red}{+4.48}          &  & 48.24\tiny\textcolor{red}{+5.16}          & 96.96\tiny\textcolor{red}{+2.47}          & 33.09\tiny\textcolor{red}{+2.64}          \\ 
\hline
BGMix                                             & 58.38& 82.60                                & 38.70                                &  & 51.98                                & 93.52                                & 38.12                                \\
\rowcolor[rgb]{0.925,0.925,0.925} \textbf{+DISep} & 63.54\tiny\textcolor{red}{+5.16}& 89.74\tiny\textcolor{red}{+7.14}          & 42.49\tiny\textcolor{red}{+3.79}          &  & 55.24\tiny\textcolor{red}{+3.26}          & 95.61\tiny\textcolor{red}{+2.09}          & 42.82\tiny\textcolor{red}{+4.70}          \\ 
\hline
TransWCD                                          & 65.53& 95.17& 49.36                                &  & 60.08                                & 95.56                                & 42.94                                \\
\rowcolor[rgb]{0.925,0.925,0.925} \textbf{+DISep} & \textbf{70.21}\tiny\textcolor{red}{+4.68}& 95.43\tiny\textcolor{red}{+0.26}          & \textbf{55.95}\tiny\textcolor{red}{+6.59} &  & \textbf{66.35}\tiny\textcolor{red}{+6.27} & 96.99\tiny\textcolor{red}{+1.43}          & \textbf{49.38}\tiny\textcolor{red}{+6.44} \\ 
\hline
MSCAM                                             & 65.03                                & 94.40                                & 50.04                                &  & 61.24                                & 94.72                                & 43.07                                \\
\rowcolor[rgb]{0.925,0.925,0.925} \textbf{+DISep} & 69.32\tiny\textcolor{red}{+4.29}          & 95.00\tiny\textcolor{red}{+0.60}          & 54.30\tiny\textcolor{red}{+4.26}          &  & 66.02\tiny\textcolor{red}{+4.78}          & \textbf{98.53}\tiny\textcolor{red}{+3.81} & 47.43\tiny\textcolor{red}{+4.36}          \\ 
\hline
MS-Former                                         & 67.31                                & 94.21                                & 53.21                                &  & 61.44                                & 94.63                                & 41.89                                \\
\rowcolor[rgb]{0.925,0.925,0.925} \textbf{+DISep} & 69.98\tiny\textcolor{red}{+2.67}          & \textbf{96.46}\tiny\textcolor{red}{+2.25}& 55.26\tiny\textcolor{red}{+2.05}          &  & 65.41\tiny\textcolor{red}{+3.97}          & 97.53\tiny\textcolor{red}{+2.90}          & 45.33\tiny\textcolor{red}{+3.44}          \\
\hline\hline
\end{tabular}
\end{table*}

\begin{table*}
\setlength{\tabcolsep}{3pt}
\renewcommand{\arraystretch}{1.2}
\centering
\caption{Quantitative improvements over baselines and comparisons with weakly-supervised change detection methods on land-cover change detection datasets. F1 score (\%), OA (\%), and IoU (\%) are reported. Red text indicates improvements brought by our DISep method, and bold font indicates the highest performance values achieved.}
\label{WSCD_m}
\begin{tabular}{ccccccccclccc} 
\hline\hline
\multirow{2}{*}{\textbf{Method}} &  & \multicolumn{3}{c}{\textbf{\textbf{DSIFN-CD}}
}                                                                                                              &  & \multicolumn{3}{c}{\textbf{SYSU-CD}} &  &\multicolumn{3}{c}{\textbf{CDD}
}\\ 
\cline{3-5}\cline{7-9}\cline{11-13}
                                 &  & \textbf{F1}                          & \textbf{\textbf{OA}}                 & \textbf{\textbf{IoU}}                 
&  & \textbf{F1}                                   & \textbf{OA}                                   & \textbf{IoU}                                    &  &\textbf{F1}                                   & \textbf{OA}                                   &\textbf{IoU}                                  
\\ 
\cline{1-1}\cline{3-5}\cline{7-9}\cline{11-13}
Kalita \textit{et al.}           &  & 23.55                                & 72.65                                & 16.22                                 
&  & 27.19                                         & 84.36                                         & 18.02                                           &  &19.60                                         & 69.66                                         &15.48                                         
\\
\rowcolor[rgb]{0.925,0.925,0.925}\textbf{+DISep}                  &  & 26.54\tiny\textcolor{red}{+3.01}          & 75.28\tiny\textcolor{red}{+2.63}          & 21.51\tiny\textcolor{red}{+5.29}           
&  & 31.82\tiny\textcolor{red}{+4.63}          & 87.45\tiny\textcolor{red}{+3.09}          & 22.73\tiny\textcolor{red}{+4.71}            &  &23.60\tiny\textcolor{red}{+4.00}          & 73.56\tiny\textcolor{red}{+3.90}          &19.62\tiny\textcolor{red}{+4.14}          
\\ 
\hline
WCDNet                           &  & 29.77                                & 72.59                                & 17.26                                 
&  & 29.68                                         & 87.09                                         & 19.11                                           &  &30.18                                         & 71.78                                         &18.36                                         
\\
\rowcolor[rgb]{0.925,0.925,0.925}\textbf{+DISep}                  &  & 35.33\tiny\textcolor{red}{+5.56}          & 76.18\tiny\textcolor{red}{+3.59}          & 25.04\tiny\textcolor{red}{+7.78}           
&  & 34.90\tiny\textcolor{red}{+5.22}          & 89.70\tiny\textcolor{red}{+2.61}          & 24.89\tiny\textcolor{red}{+5.78}            &  &35.40\tiny\textcolor{red}{+5.22}          & 75.96\tiny\textcolor{red}{+4.18}          &23.86\tiny\textcolor{red}{+5.50}          
\\ 
\hline
FCD-GAN                          &  & 40.26                                & 73.27                                & 29.94                                 
&  & 44.23                                         & 84.58                                         & 31.69                                           &  &39.23                                         & 72.04                                         &30.55                                         
\\
\rowcolor[rgb]{0.925,0.925,0.925}\textbf{+DISep}                  &  & 45.42\tiny\textcolor{red}{+5.16}          & 76.74\tiny\textcolor{red}{+3.47}          & 34.83\tiny\textcolor{red}{+4.89}           
&  & 49.80\tiny\textcolor{red}{+5.57}          & 87.15\tiny\textcolor{red}{+2.57}          & 36.90\tiny\textcolor{red}{+5.21}            &  &42.84\tiny\textcolor{red}{+3.61}          & 75.43\tiny\textcolor{red}{+3.39}          &33.02\tiny\textcolor{red}{+2.47}          
\\ 
\hline
BGMix                            &  & 45.80                                & 77.96                                & 31.94                                 
&  & 52.20                                         & 82.71                                         & 37.42                                           &  &45.36                                         & 78.04                                         &32.12                                         
\\
\rowcolor[rgb]{0.925,0.925,0.925}\textbf{+DISep}                  &  & 50.02\tiny\textcolor{red}{+4.22}          & 84.00\tiny\textcolor{red}{+6.04}          & 35.69\tiny\textcolor{red}{+3.75}           
&  & 57.14\tiny\textcolor{red}{+4.94}          & 85.40\tiny\textcolor{red}{+2.69}          & 41.69\tiny\textcolor{red}{+4.27}            &  &49.52\tiny\textcolor{red}{+4.16}          & 82.51\tiny\textcolor{red}{+4.47}          &35.94\tiny\textcolor{red}{+3.82}          
\\ 
\hline
TransWCD                         &  & 53.41                                & 83.05                                & 36.44                                 
&  & 60.72                                         & 84.01                                         & 41.88                                           &  &51.38                                         & 88.04                                         &32.55                                         
\\
\rowcolor[rgb]{0.925,0.925,0.925}\textbf{+DISep}                  &  & \textbf{57.78}\tiny\textcolor{red}{+4.37} & 83.36\tiny\textcolor{red}{+0.31}          & 41.25\tiny\textcolor{red}{+4.81}           
&  & \textbf{65.93}\tiny\textcolor{red}{+5.21}& 86.47\tiny\textcolor{red}{+2.46}& 46.30\tiny\textcolor{red}{+4.42}&  &\textbf{54.70}\tiny\textcolor{red}{+3.32}& \textbf{89.71}\tiny\textcolor{red}{+1.67}&\textbf{37.64}\tiny\textcolor{red}{+5.09}\\ 
\hline
MSCAM                            &  & 52.79                                & 84.11                                & 37.38                                 
&  & 60.54                                         & 85.35                                         & 41.16                                           &  &51.00                                         & 83.97                                         &32.13                                         
\\
\rowcolor[rgb]{0.925,0.925,0.925}\textbf{+DISep}                  &  & 57.45\tiny\textcolor{red}{+4.66}          & \textbf{87.31}\tiny\textcolor{red}{+3.20} & \textbf{41.68}\tiny\textcolor{red}{+4.30}  
&  & 65.32\tiny\textcolor{red}{+4.78}          & \textbf{87.52}\tiny\textcolor{red}{+2.17}          & 45.73\tiny\textcolor{red}{+4.57}            &  &55.38\tiny\textcolor{red}{+4.38}          & 87.83\tiny\textcolor{red}{+3.86}          &36.22\tiny\textcolor{red}{+4.09}          
\\ 
\hline
MS-Former                        &  & 52.55                                & 83.32                                & 37.04                                 
&  & 59.61                                         & 83.72                                         & 41.52                                           &  &51.67                                         & 83.49                                         &31.82                                         
\\
\rowcolor[rgb]{0.925,0.925,0.925}\textbf{+DISep}                  &  & 56.38\tiny\textcolor{red}{{+3.83}}& 86.54\tiny\textcolor{red}{+3.22}          & 41.40\tiny\textcolor{red}{+4.36}           &  & 64.75\tiny\textcolor{red}{+5.14}          & 85.82\tiny\textcolor{red}{+2.10}          & \textbf{46.93}\tiny\textcolor{red}{+5.41}            &  &54.93\tiny\textcolor{red}{+3.26}          & 87.90\tiny\textcolor{red}{+4.41}          &35.98\tiny\textcolor{red}{+4.16}          \\
\hline\hline
\end{tabular}
\end{table*}

\begin{figure*}[!ht]
	\centering
	\includegraphics[scale=0.85]{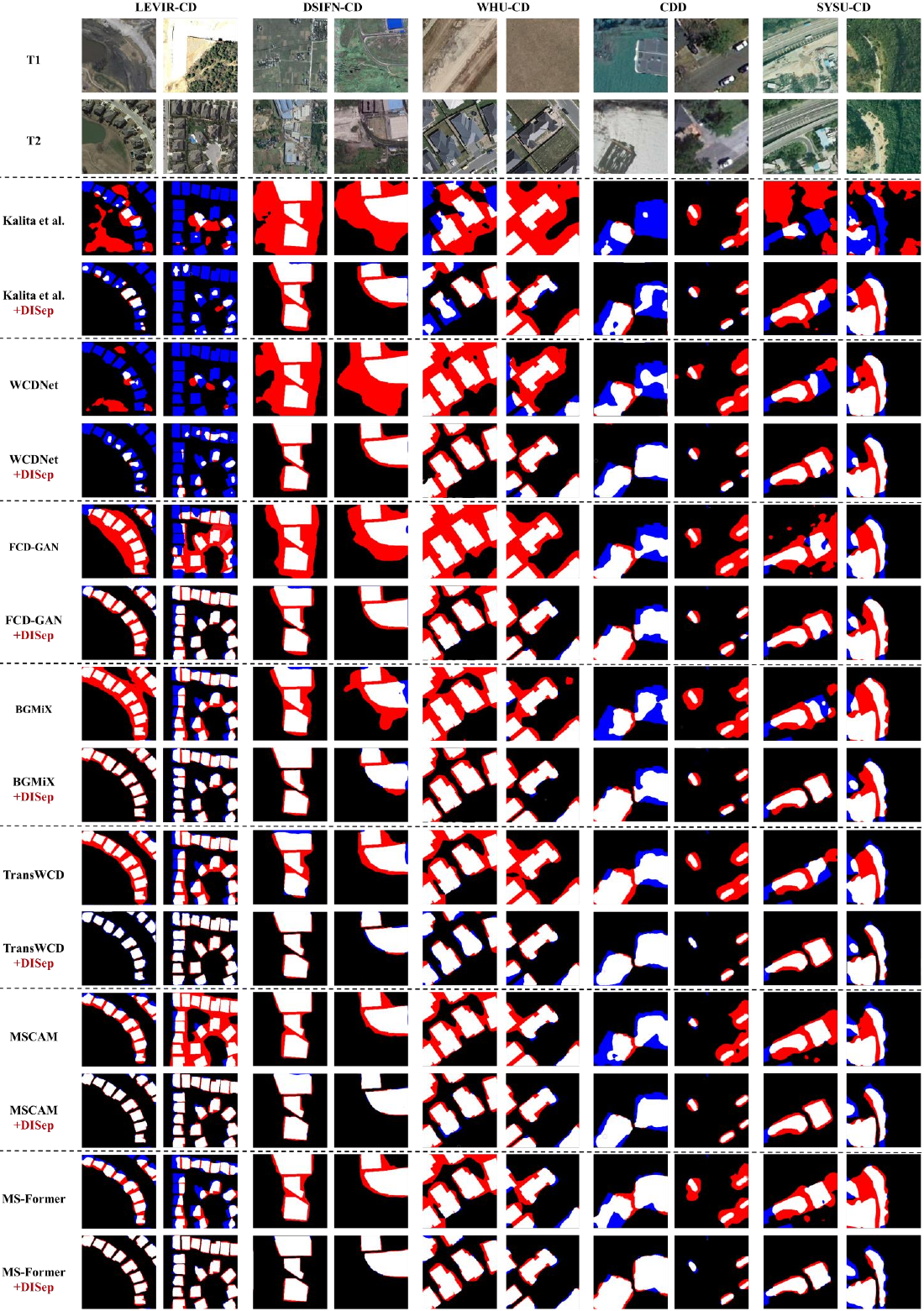}
	\caption{Qualitative improvements of our DISep. For clarity, in the predictions, \textcolor{red}{false positive (erroneous changed) pixels} are marked in red color, and \textcolor{blue}{false negative (erroneous unchanged) pixels} are marked in blue color..}	\label{experiment1}
\end{figure*}

We evaluate DISep on top of existing WSCD methods, including WCDNet, FCD-GAN, BGMix, Kalita \textit{et al.}'s approach, TransWCD, MSCAM, and MS-Former, on building change detection datasets (WHU-CD and LEVIR-CD), as shown in Table \ref{WSCD_b}, and on land-cover change detection datasets (DSIFN-CD, CDD, and SYSU-CD), as shown in Table \ref{WSCD_m}. The complete qualitative results are presented in Figure \ref{experiment1}.

\subsubsection{Quantitative Results}
Integrating DISep notably boosts the performance indicators across the board on these five datasets. For instance, on the WHU-CD dataset, DISep enhances TransWCD's performance, yielding a +4.68\% increase in F1 score and a +6.59\% increase in IoU. Additionally, applying DISep to MS-Former results in a +2.67\% increase in F1 score and a significant +2.25\% increase in OA, and integrating DISep with MSCAM yields a +4.29\% improvement in F1 score. Improvements over WCDNet are even more striking, with +7.91\% F1 score and +8.24\% IoU. FCD-GAN sees an improvement of +4.11\% F1 score, indicating that DISep effectively enhances generative models for better change detection. And the integration of DISep with BGMix results in a +5.16\% rise in the F1 score. Applying DISep to Kalita \textit{et al.}'s method leads to a +4.82\% improvement in the F1 score.

On the more challenging LEVIR-CD dataset, DISep achieves gains of +6.27\% F1 score and +6.44\% IoU against TransWCD, and +4.41\% F1 score and +3.89\% IoU against WCDNet. Remarkably, DISep sets a new record with F1 scores of 66.35\% on the LEVIR-CD dataset. This superior performance of DISep is consistent across all metrics. BGMix, enhanced with DISep, also displayed impressive gains, with a +3.26\% increase in F1 score and a +2.09\% improvement in OA. Similarly, MSCAM showed an increase of +4.78\% in F1 score and +4.36\% in IoU, while MS-Former showed gains of +3.97\% in F1 score and +3.44\% in IoU.

On the DSIFN-CD dataset, the enhancements are particularly remarkable with WCDNet and FCD-GAN. When augmented with DISep, WCDNet experiences the highest increase, with a +5.56\% boost in F1 score and a +7.78\% increase in IoU, showcasing the broad applicability of DISep. FCD-GAN also sees significant improvements, with a +5.16\% increase in F1 score and a +4.89\% increase in IoU, indicating strong performance in diverse change detection scenarios. Additionally, MSCAM achieves a +4.66\% increase in F1 score and a +4.30\% improvement in IoU.

On the SYSU-CD dataset, the improvements are noteworthy. Applying DISep to Kalita \textit{et al.}'s method results in a +4.63\% F1 score and a +4.71\% IoU increase. WCDNet sees gains of +5.22\% F1 score and +5.78\% IoU, while FCD-GAN improves by +5.57\% F1 score and +5.21\% IoU. BGMix shows increases of +4.94\% in F1 score and +4.27\% in IoU. TransWCD records an impressive +5.21\% F1 score and +4.42\% IoU increase, setting new performance benchmarks. MSCAM and MS-Former also benefit significantly, with MSCAM achieving a +4.78\% F1 score and +4.57\% IoU increase, and MS-Former gaining +5.14\% F1 score and +5.41\% IoU.

On the CDD dataset, DISep delivers substantial enhancements. Kalita \textit{et al.}'s method sees a +4.00\% increase in F1 score and a +4.14\% IoU improvement. WCDNet benefits from a +5.22\% F1 score and a +5.50\% IoU increase. FCD-GAN's performance improves by +3.61\% in F1 score and +2.47\% in IoU. BGMix records gains of +4.16\% in F1 score and +3.82\% in IoU. TransWCD achieves an additional +3.32\% F1 score and +5.09\% IoU, marking significant performance improvements. MSCAM sees a +4.38\% increase in F1 score and a +4.09\% improvement in IoU, while MS-Former gains +3.26\% in F1 score and +4.16\% in IoU. It can be seen that our method is not only effective for changes on buildings but also generalizes well to land-cover datasets.

Across all five datasets, DISep has established new benchmarks, notably achieving record-breaking F1 scores of 66.35\% on LEVIR-CD, 70.21\% on WHU-CD, 57.78\% on DSIFN-CD, 65.93\% on SYSU-CD, and 54.70\% on CDD. In summary, DISep significantly enhances WSCD methods across multiple datasets, setting new performance records and improving accuracy. Its consistent use improves detection capabilities, establishing it as a critical plug-and-play tool for advancing change detection research.

\subsubsection{Qualitative Results} 
As illustrated in Figure \ref{experiment1}, we visually demonstrate the qualitative improvements achieved by DISep. Compared to the seven baseline methods, DISep effectively alleviates instance lumping in dense instance scenarios. With the aid of DISep, the models now have an improved ability to separate and accurately identify individual changes, enhancing the clarity and accuracy of the predicted results. DISep reduces \textcolor{red}{false positive pixels} (i.e., erroneous changed pixels) in the changed instance gaps, and also eliminates some \textcolor{blue}{false negative pixels} (i.e., erroneous unchanged pixels), as shown in the second line. The change predictions of DISep closely align with pixel-level ground truths, providing compelling evidence for its effectiveness in enhancing the accuracy and reliability of WSCD in instance-crowded scenes. Additionally, these hard examples from the LEVIR-CD dataset illustrate the effectiveness of our proposed method in handling complex changes and demonstrate its robustness.

\subsection{Ablation Study}
The following experimental section uses the Transformer-based TransWCD and ConvNet-based WCDNet as the default baselines for our ablation study.

\subsubsection{Computational Cost}
Our DISep is implemented in a plug-and-play mechanism. To evaluate its computational efficiency, we measure the training and inference time of our DISep compared to the TransWCD baseline. As shown in Table \ref{time}, DISep only adds 0.02 and 0.03 minutes per 100 iterations to the training time on the two baselines, respectively. Moreover, DISep is only involved in the training phase, so it adds no inference overhead. The GPU memory overhead is minimal, increasing by only about 0.2GB for both ConvNet-based and Transformer-based methods.
\begin{table}[h]
\renewcommand{\arraystretch}{1.2}
\centering
\caption{Training time (minutes per 100 iterations) and GPU overhead (GB) of DISep. All results are evaluated on the training split of the LEVIR-CD dataset.}
\label{time}\small
\begin{tabular}{ccc} 
\hline\hline
\textbf{Method}           & \textbf{Training Time} & \textbf{GPU Overhead}  \\ 
\hline
WCDNet                    & 2.69                   & 6.32                   \\
\textbf{Ours w/ WCDNet}   & \textbf{2.71}          & \textbf{6.34}          \\ 
\hline
TransWCD                  & 3.46                   & 4.52                   \\
\textbf{Ours w/ TransWCD} & \textbf{3.49}          & \textbf{4.54}          \\
\hline\hline
\end{tabular}
\end{table}

\subsubsection{Weight Factor of \texorpdfstring{$\mathcal{L}_{sep}$}{L_sep}}
\begin{figure*}
\centering
\includegraphics[scale=0.50]{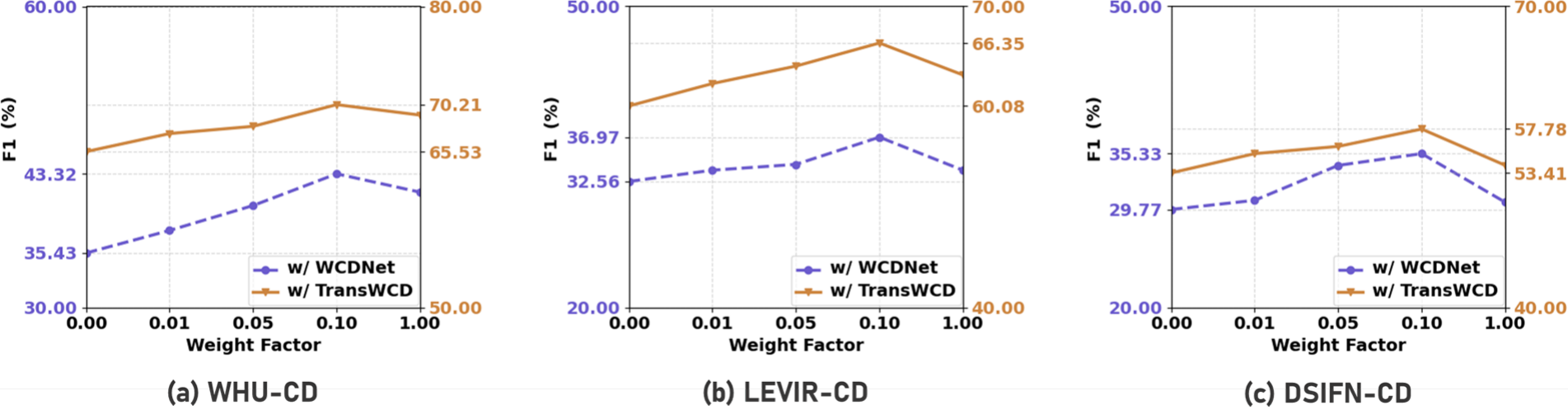}
	\caption{Weight factor of $\mathcal{L}_{sep}$. We vary the weight factor of our separation loss on the overall loss, on top of the WCDNet and TransWCD baselines. F1 score (\%) is reported on the WHU-CD, LEVIR-CD, and DSIFN-CD datasets.}

	\label{experiment3}
\end{figure*}
We investigate the sensitivity of varying the weight factor of our separation loss on the overall loss, as shown in Figure \ref{experiment3}. Across the range from $0$ to $0.1$, we observe consistent enhancements in performance. Notably, on the WHU-CD dataset, setting the weight factor to $0.1$ brings the most significant gains, with WCDNet's F1 score increasing by 4.68\% and TransWCD peaking at 70.21\% F1 score. Beyond this optimal point, further increasing the weight factor may decrease results. This negative effect arises because there is an excessive focus on pixel clustering rather than the primary classification loss, disrupting the balance in the overall loss function. A similar trend is observed in the LEVIR-CD and DSIFN-CD datasets, where we also find that the best F1 score occurs when the weight is set to 0.1.

\subsubsection{Different High- and Low-Pass Thresholds \texorpdfstring{$T_{h}$ and $T_{l}$}{T\_h and T\_l}}
As shown in Table \ref{high_pass}, we assess the impact of the high- and low-pass thresholds $T_{h}$ and $T_{l}$ on instance localization. Our analysis reveals an optimal threshold combination of $T_{h} = 0.60$ and $T_{l} = 0.40$, which achieves the highest F1 score and Intersection over Union (IoU) among the values tested. This combination demonstrates the most effective balance between minimizing Type-I errors (i.e., false positives), optimizing the pixel sampling size, and setting both the high- and low-pass thresholds to a CAM score of 0.45, which results in performance that is not only below the baseline but markedly worse. This decline is primarily attributed to the significant increase in false positive pixels. Similar results are observed on other datasets.
\begin{table}[h]
\centering
\renewcommand{\arraystretch}{1.2}
\caption{Different choices for high--pass thresholds \textbf{$\boldsymbol{T_{h}}$} and low--pass thresholds \textbf{$\boldsymbol{T_{l}}$}. F1 score (\%), OA (\%), and IoU (\%) are reported on the LEVIR-CD dataset. Bold font indicates the highest performance values achieved.}
\label{high_pass}\small
\begin{tabular}{cc|ccc} 
\hline\hline
\textbf{$\boldsymbol{T_{h}}$} & \textbf{$\boldsymbol{T_{l}}$}  & \textbf{F1}    & \textbf{OA}    & \textbf{IoU}   \\ 
\hline
Baseline                      & Baseline                       & 60.08& 95.56& 42.94\\ 
\hline
0.65                          & \multirow{5}{*}{0.45}          & 63.01          & 95.89          & 48.25           \\
0.60                          &                                & 63.33          & 95.78          & 48.27           \\
0.55                          &                                & 62.54          & 95.42          & 46.51           \\
0.50                          &                                & 61.90          & 95.35          & 46.23           \\
0.45                          &                                & 58.72& 95.04& 40.27\\ 
\hline
0.65                          & \multirow{5}{*}{\textbf{0.40}} & 64.94& 95.86          & 48.45\\
\textbf{0.60}                 &                                & \textbf{66.35} & \textbf{96.99} & \textbf{49.38}  \\
0.55                          &                                & 65.45          & 97.10          & 48.20           \\
0.50                          &                                & 59.87          & 96.93          & 47.56           \\
0.45                          &                                & 58.93          & 96.99          & 45.61           \\ 
\hline
0.65                          & \multirow{5}{*}{0.35}          & 65.49& 96.09          & 48.79\\
0.60                          &                                & 64.74          & 95.78          & 46.31           \\
0.45                          &                                & 61.90          & 95.63          & 48.08           \\
0.50                          &                                & 59.11          & 95.78          & 45.44           \\
0.55                          &                                & 58.20          & 95.45          & 43.67           \\

\hline\hline
\end{tabular}
\end{table}

\subsubsection{Different Separation Loss Functions}\label{sec:diff_loss}
The separation loss defines the learning approach for clustering intra-instance pixels. This section evaluates the performance of different objective functions applied to our separation loss. These include the commonly used contrastive loss \citep{hjelm2019learning}, triplet loss \citep{schroff2015facenet}, cross-temporal loss prevalent in fully-supervised change detection \citep{SCPFCD,du2022weakly}, and centroid-to-pixel contrast adapted from \citep{du2022weakly}. Our scheme focuses on pixel-to-centroid clustering. Across all these methods, we treat pixels as units of instances rather than as categories.

As shown in Table \ref{pixel_loss}, our pixel-to-centroid clustering approach markedly outperforms the triplet, contrastive, and cross-temporal losses, achieving F1 score improvements of +2.60\%, +2.30\%, and +3.67\%, respectively. It's important to note that the pixel-to-centroid contrast approach includes an additional centroid-to-centroid contrast added to our pixel-to-centroid clustering. While this contrast aims to enforce inter-instance discriminability, it unfortunately reduces the F1 score by -2.10\%. We argue that the reason might be that this centroid-to-centroid contrast leads to excessive dissimilarity among changed pixels from different instances, which in turn hampers the discrimination between changed and unchanged pixels.

\begin{table}
\centering
\renewcommand{\arraystretch}{1.2}
\caption{Different objective functions for the separation loss. F1 score (\%), OA (\%), and IoU (\%) are reported on the LEVIR-CD dataset. Bold font indicates the highest performance values achieved.}
\label{pixel_loss}\small
\begin{tabular}{cccc} 
\hline\hline
\textbf{Separation Loss}& \textbf{F1} & \textbf{OA} & \textbf{IoU}  \\ 
\hline
Baseline                                 & 60.08 & 95.56 & 42.94 \\ 
\hline
+Triplet                                 & 63.75 & 96.80 & 46.44 \\
+Contrastive                             & 64.05 & 96.85 & 46.83 \\
\multicolumn{1}{c}{+Cross-Temporal}      & 62.68 & 96.70 & 45.62 \\
\textbf{ +Pixel-to-Centroid Clustering  }         & \textbf{66.35} & \textbf{96.99} & \textbf{49.38}  \\
+Pixel-to-Centroid Contrast              & 64.25 & 96.62 & 49.38 \\
\hline\hline
\end{tabular}
\end{table}

\begin{figure*}[!ht]
	\centering
	\includegraphics[scale=0.55]{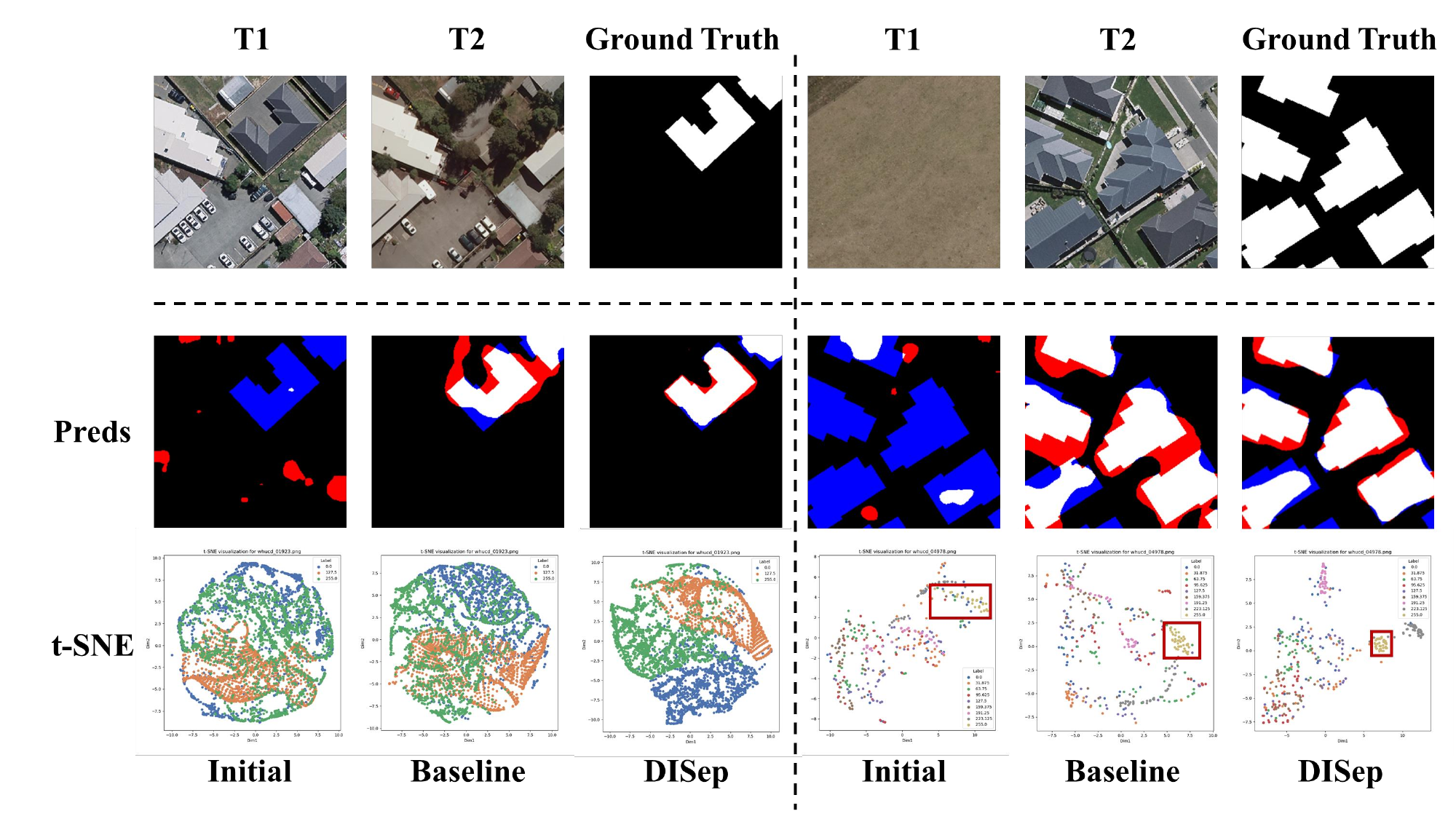}
 \caption{Evolution of instance-wise feature distribution. Notable enhancements are observed in scenarios of different instance densities. For clarity, the third row marks pixels from different instances, including unchanged pixels, in various colors. For ease of observation, the number of sampling points is determined based on the size of the smallest instance in the image examples.}
 \label{experiment4}
\end{figure*}

\subsubsection{Different Instance Sampling Scopes}
We assess different instance sampling scopes, which vary from considering only \textit{changed-in-changed} pixels, adding \textit{unchanged-in-changed} pixels, and also \textit{unchanged-in-unchanged} pixels. The performance impact of these varied scopes is detailed in Table \ref{sampling_scope}.

Focusing solely on pixels within changed instances (CC) leads to a +4.90\% increase in the F1 score and a +4.70\% rise in IoU. Expanding this scope to include both unchanged and changed pixels within changed images (CC+CU), results in further enhancements, with an additional +0.90\% in IoU and +0.63\% in the F1 score. Lastly, extending the calculation to unchanged pixels in unchanged images (CC+CU+UU) provides an extra increase of +0.62\% in IoU and +0.47\% in F1 score.
\begin{table}[!ht]
\renewcommand{\arraystretch}{1.5}
\centering
\caption{Different instance sampling scopes. Performance is evaluated by only calculating the changed-in-changed instances (CC), adding unchanged-in-changed instances (CU), and further adding unchanged-in-unchanged instances (UU). F1 score (\%), OA (\%), and IoU (\%) are reported on the LEVIR-CD dataset. Bold font indicates the highest performance values achieved.}
\label{sampling_scope}\small
\begin{tabular}{cccc} 
\hline\hline
\textbf{Sampling Scope}& \textbf{F1} & \textbf{OA} & \textbf{IoU}  \\ 
\hline
Baseline                  & 60.08 & 95.56 & 42.94 \\ 
\hline
+CC                       & 64.98       & 96.64       & 47.64         \\
+CC+CU                    & 65.88       & 96.76       & 48.27\\
\textbf{+CC+CU+UU}                 & \textbf{66.35} &\textbf{96.99} & \textbf{49.38}  \\
\hline\hline
\end{tabular}
\end{table}

\section{Discussion}
\subsection{Instance Manifold Visualization}
To track the evolution of pixel feature grouping, we compare the final feature manifold of our DISep and the baseline method (TransWCD) using t-SNE \citep{maaten2008visualizing}. As depicted in Figure \ref{experiment4}, we selected multiple images with varying instance densities for comprehensive evaluation. In the third row of this figure, we observe that pixels from the same instance gradually cluster more closely. In contrast, overlapping distributions between different instances progressively decrease, significantly reducing the overall distribution uncertainty.

\begin{figure}[h]
	\centering
	\includegraphics[scale=0.68]{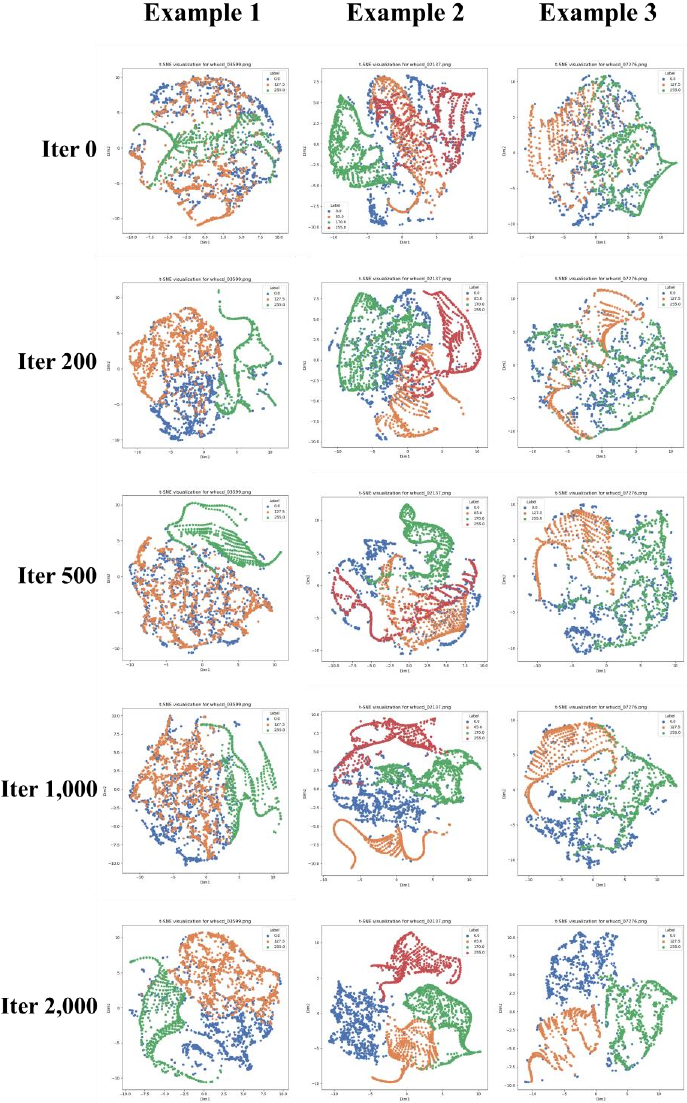}
	\caption{Detailed evolution in distribution of instance-wise pixel features.}
\label{appendix2}
\end{figure}

The second row shows the evolution of change prediction. Moreover, the examples on the left side illustrate that DISep also enhances single-instance predictions, with the predictions more closely matching the ground truths. This progression validates the effectiveness and universality of DISep in WSCD scenarios with various instance densities. Additionally, Figure \ref{appendix2} provides a detailed view of the evolution in instance-wise feature distributions, showing the increasing separability of features with the training iterations.
\subsection{Comparison with Other Auxiliary Modules}

\begin{table}
\centering
\renewcommand{\arraystretch}{1.2}
\caption{Comparison with other auxiliary modules on the LEVIR-CD datasets. F1 score (\%), OA (\%), and IoU (\%) are reported. Red tiny text indicates the improvements brought by our DISep method, and bold font indicates the highest performance values achieved.}
\label{auxiliary}
\begin{tabular}{cccc} 
\hline\hline
\multirow{2}{*}{\textbf{Method}} & \multicolumn{3}{c}{\textbf{LEVIR-CD}}               \\ 
\cline{2-4}
                                 & \textbf{F1}  & \textbf{OA}  & \textbf{IoU}  \\ 
\hline
Baseline                         & 60.08        & 95.56        & 42.94         \\
\hline
+DenseCRF                        & 61.63\tiny\textcolor{red}{+1.55}& 96.01\tiny\textcolor{red}{+0.45} & 43.90\tiny\textcolor{red}{+0.96} \\
+Affinity Learning               & 61.10\tiny\textcolor{red}{+1.02} & 96.18\tiny\textcolor{red}{+0.62} & 43.50\tiny\textcolor{red}{+0.56} \\
\textbf{+DISep}& \textbf{66.35}\tiny\textcolor{red}{+6.27} & \textbf{96.99}\tiny\textcolor{red}{+1.43} & \textbf{49.38}\tiny\textcolor{red}{+6.44} \\ 
\hline\hline
\end{tabular}
\end{table}

We further compare our method with other auxiliary modules to improve instance predictions under image-level weakly-supervised change detection, including DenseCRF \citep{krahenbuhl2011efficient} and Affinity Learning \citep{xu2021leveraging}. DenseCRF uses Gaussian edge potentials for efficient inference in fully connected Conditional Random Fields, which is popular for post-processing in weakly-supervised semantic segmentation. Affinity Learning refines coarse predictions by learning pixel affinity matrices based on feature similarities, effectively capturing the intricate structures within the image. Both methods have proven effective in weakly-supervised semantic segmentation \citep{liu2019visual}.

In this section, we evaluate the instance prediction improvements of these methods for WSCD, highlighting the superiority of our DISep. Specifically, we compare these auxiliary modules using the TransWCD baseline on the LEVIR-CD dataset, as shown in Table \ref{auxiliary}. Compared to DenseCRF, the DISep method shows significant improvements in all metrics. Specifically, DISep improves the F1 score by +3.72\% (66.35\% vs. 61.63\%), OA by +0.98\% (96.99\% vs. 96.01\%), and IoU by +5.48\% (49.38\% vs. 43.90\%). Similarly, when compared to Affinity Learning, DISep demonstrates notable enhancements with an F1 score improvement of +5.25\% (66.35\% vs. 61.10\%), OA improvement of +0.81\% (96.99\% vs. 96.18\%), and IoU improvement of +5.88\% (49.38\% vs. 43.50\%).

These suboptimal improvements of DenseCRF and Affinity Learning are attributed to the fact that these commonly used weakly-supervised semantic segmentation methods primarily rely on the pixel relationships within the original images. In the case of paired image inputs in WSCD, only the inaccurate pixel relationships in the bi-temporal difference images can be used, where adjacent pixel relationships may lack coherence and smoothness. As a result, these methods, which perform well in weakly-supervised semantic segmentation, are less effective in WSCD. Our tailored DISep method is more suitable for WSCD, as it is specifically designed to handle the unique challenges posed by this task.

\subsection{Extending to Fully-Supervised CD}
Considering that current fully-supervised change detection (FSCD) methods sometimes encounter the same issue of change lumping, we extend our DISep approach to FSCD. 
\begin{table}
\centering
\renewcommand{\arraystretch}{1.2}
\caption{Extension of our DISep on fully-supervised change detection. F1 score (\%),  OA (\%), and IoU (\%) are reported on the LEVIR-CD dataset.}
\label{FSSS}\small
\begin{tabular}{cccc} 
\hline\hline
\multirow{2}{*}{\textbf{\textbf{Fully-Supervised~}Method}} & \multicolumn{3}{c}{\textbf{LEVIR-CD}}               \\ 
\cline{2-4}
                                                           & \textbf{F1}  & \textbf{OA}  & \textbf{IoU}  \\ 
\hline
SNUNet                                                     & 91.88& 98.50& 85.84\\
\textbf{Ours w/ SNUNet}                                    & \textbf{93.14}& \textbf{99.09}& \textbf{86.29}\\
CTD-Former& 92.71& 98.62& 87.11\\
\textbf{Ours w/ CTD-Former}& \textbf{94.03}& \textbf{99.07}& \textbf{88.34}\\
\hline\hline
\end{tabular}
\end{table}
Within the fully-supervised change detection (FSCD) framework, we utilize the binary pixel-level ground truths $\mathcal{Y}$ to generate precise instance localization. This involves directly mapping elements in $\mathcal{Y}$ to create the changed instance localization maps $\mathcal{M}_c$ and the unchanged instance localization maps $\mathcal{M}_{uc}$ as follows:

\[
\mathcal{M}_c(i,j) = 
\begin{cases} 
1 & \text{if } \mathcal{Y}(i,j) = 1, \\
0 & \text{otherwise},
\end{cases}
\]
and
\[
\mathcal{M}_{uc}(i,j) = 
\begin{cases} 
1 & \text{if } \mathcal{Y}(i,j) = 0, \\
0 & \text{otherwise}.
\end{cases}
\]

where $\mathcal{M}_c$ accurately reflects pixels identified as changed (i.e., where $\mathcal{Y}^i = 1$), and $\mathcal{M}_{uc}$ correctly represents pixels that remain unchanged (i.e., where $\mathcal{Y}^i = 0$), effectively inverting the values for unchanged pixels in $\mathcal{M}_{uc}$. The processes for subsequent instance retrieval and separation are the same as described in WSCD.

We evaluate DISep's performance in FSCD using two well-known FSCD baseline methods: SNUNet \citep{37fang2022snunet} and CTD-Former \citep{Zhang2023relation}. As shown in Table \ref{FSSS}, DISep consistently demonstrates performance improvements in FSCD scenarios. When integrated with SNUNet, DISep achieves increases of +1.26\% F1 score, +0.59\% OA, and +0.45\% IoU. Similarly, when combined with CTD-Former, DISep shows gains of +1.32\% F1 score, +0.45\% OA, and +1.23\% IoU. These results confirm the effectiveness of DISep and highlight the benefits of incorporating instance-wise contextual information for pixels in change detection tasks.

\section{Conclusion}

In weakly-supervised change detection, unchanged pixels between changed instances are often misclassified as changed, leading to merged instances in change predictions. To address this issue, we propose a plug-and-play method, DISep, to separate dense instances. Initially, we perform instance localization with a high-pass threshold to obtain instance localization masks from CAMs. Then, we implement an innovative instance retrieval process using a straightforward connectivity search to acquire instance identity masks. Finally, we introduce a separation loss to guide the clustering of intra-instance pixels in the embedding features, based on the instance identity masks.

Our evaluation of DISep on seven existing ConvNet-based and Transformer-based WSCD methods shows significant improvements on the WHU-CD, LEVIR-CD, DSIFN-CD, SYSU-CD, and CDD datasets. Further experiments indicate that DISep can also enhance fully-supervised change detection tasks, demonstrating its effectiveness and versatility.




{
\bibliographystyle{elsarticle-harv} 
\bibliography{reference}
}

\end{document}